%% file: main.tex
\documentclass[10pt,twocolumn,letterpaper]{article}

\usepackage[pagenumbers]{cvpr} 

\input{preamble}

\definecolor{cvprblue}{rgb}{0.21,0.49,0.74}
\usepackage[pagebackref,breaklinks,colorlinks,citecolor=cvprblue]{hyperref}

\title{Range-EDIT: Semantic Mask Guided LiDAR Scene Editing}

\author{Suchetan G. Uppur\\
GeoAI4Cities Lab, IISER Bhopal, India\\
{\tt\small suchetan21@iiserb.ac.in}
\and
Hemant Kumar\\
GeoAI4Cities Lab, IISER Bhopal, India\\
{\tt\small hemant23@iiserb.ac.in}
\and
Vaibhav Kumar\\
GeoAI4Cities Lab, IISER Bhopal, India\\
{\tt\small vaibhav@iiserb.ac.in}
}

\begin{document}
\maketitle
\input{sec/abstract}

\input{sec/introduction}
\input{sec/related_work}

\input{sec/method}

\input{sec/experiments}

\input{sec/conclusion}
{
    \small
    \bibliographystyle{ieeenat_fullname}
    \bibliography{main}
}

\end{document}

%% file: preamble.tex
\usepackage[dvipsnames]{xcolor}


%% file: sec/abstract.tex
\begin{abstract}
Training autonomous driving and navigation systems requires large and diverse point cloud datasets that capture complex edge case scenarios from various dynamic urban settings. Acquiring such diverse scenarios from real-world point cloud data, especially for critical edge cases, is challenging, which restricts system generalization and robustness. Current methods rely on simulating point cloud data within handcrafted 3D virtual environments, which is time-consuming, computationally expensive, and often fails to fully capture the complexity of real-world scenes. To address some of these issues, this research proposes a novel approach that addresses the problem discussed by editing real-world LiDAR scans using semantic mask-based guidance to generate novel synthetic LiDAR point clouds.
\\We incorporate range image projection and semantic mask conditioning to achieve diffusion-based generation. Point clouds are transformed to 2D range view images, which are used as an intermediate representation to enable semantic editing using convex hull-based semantic masks. These masks guide the generation process by providing information on the dimensions, orientations, and locations of objects in the real environment, ensuring geometric consistency and realism. This approach demonstrates high-quality LiDAR point cloud generation, capable of producing complex edge cases and dynamic scenes, as validated on the KITTI-360 dataset. This offers a cost-effective and scalable solution for generating diverse LiDAR data, a step toward improving the robustness of autonomous driving systems.
\end{abstract}

%% file: sec/introduction.tex
\section{Introduction}
\label{sec:introduction}
LiDAR (Light Detection And Ranging) is a key component in autonomous vehicle systems \cite{shi2020pv, shi2019pointrcnn}, which provides accurate and real-time 3D environmental mapping through the points. This technology provides autonomous systems to accurately perceive 3D environments, surrounding features and obstacles, and dynamic objects \cite{li2020deeplearninglidarpoint, Li_2020} and present its importance for the system. To effectively train autonomous systems for the real-world environment, large and diverse datasets of LiDAR are needed. These datasets must contain a different range of scenarios, including weather conditions, urban and rural settings, and dynamic interactions with other vehicles, pedestrians, and infrastructure, ensuring that the system can generalize and perform reliably across all possible real-world conditions \cite{geiger2012we,caesar2020nuscenes}. Filling this gap, real-world scans of LiDAR are needed, but it is challenging to acquire, time-consuming, and requires accurate registration. Due to this, it fails to capture critical edge cases, resulting in limited data diversity that reduces the robustness of autonomous systems. The scarcity of such edge cases in real-world datasets can lead to insufficient training for learning models, reducing their ability to generalize effectively across real-world environments of such diverse scenarios \cite{li2020deeplearninglidarpoint,zhou20213dshapegenerationcompletion}. Currently, the optimal method of LiDAR data generation involves simulating LiDAR data in a hand-crafted 3D simulation world \cite{manivasagam2020lidarsimrealisticlidarsimulation}. Although effective, this method is time-consuming, requires high computational resources, scene modeling, and expertise in mapping real-world scenarios and sensor behaviors. Furthermore, simulated environments may not fully capture the nuances of real-world conditions, potentially introducing artifacts into the data.
\\We propose an alternative approach to address this problem by introducing a method for minimally editing real-world LiDAR scans to generate the required edge cases, thereby enhancing data diversity while keeping the editing process straightforward and cost-effective.\\
We make use of range image projection \cite{li2016vehicledetection3dlidar} to enable the transformation of data representation between LiDAR point clouds in continuous 3D space and range images in pixel space. Operating in pixel space rather than continuous 3D space provided advantages in terms of computational complexity, as it allows avoidance of operations over the largely empty regions typically present in LiDAR point clouds, an issue observed in some previous works on LiDAR generation \cite{xiong2023ultralidarlearningcompactrepresentations}.
\\Diffusion-based models have significantly increased both the quality and efficiency of image generation \cite{zhang2023surveydiffusionbasedimage,rombach2022highresolutionimagesynthesislatent}, and previous studies demonstrated the effectiveness of the diffusion model in range image generation \cite{hu2024rangeldmfastrealisticlidar,ran2024realisticscenegenerationlidar}. This method also leverages semantic masks as a condition during the generation process; these masks provide critical information regarding the shape, orientation, and position of the object to be generated within the scene, thereby addressing the localization problem. Additionally, a region-focused loss was employed to simplify the training objective and guide the generation process toward the region of interest within range-view images. Our contributions can be summarized as follows:
\begin{itemize}
    \item To the best of our knowledge, this is the first attempt to perform object-level semantic editing of LiDAR point clouds without reliance on LiDAR scan simulation or pre-defined object 3-D models.
    \item To the best of our knowledge, there is no prior work done to perform diffusion-based semantic image editing on range-view images.
    \item We explore convex hull masking as a means of edit localization.
    \item We establish a baseline for future semantic editing work on the KITTI-360 dataset, as the first work of its kind.
\end{itemize}

%% file: sec/related_work.tex
\section{Related Work}
\label{sec:related_work}

\begin{figure*}[h]
    \includegraphics[width=\textwidth]{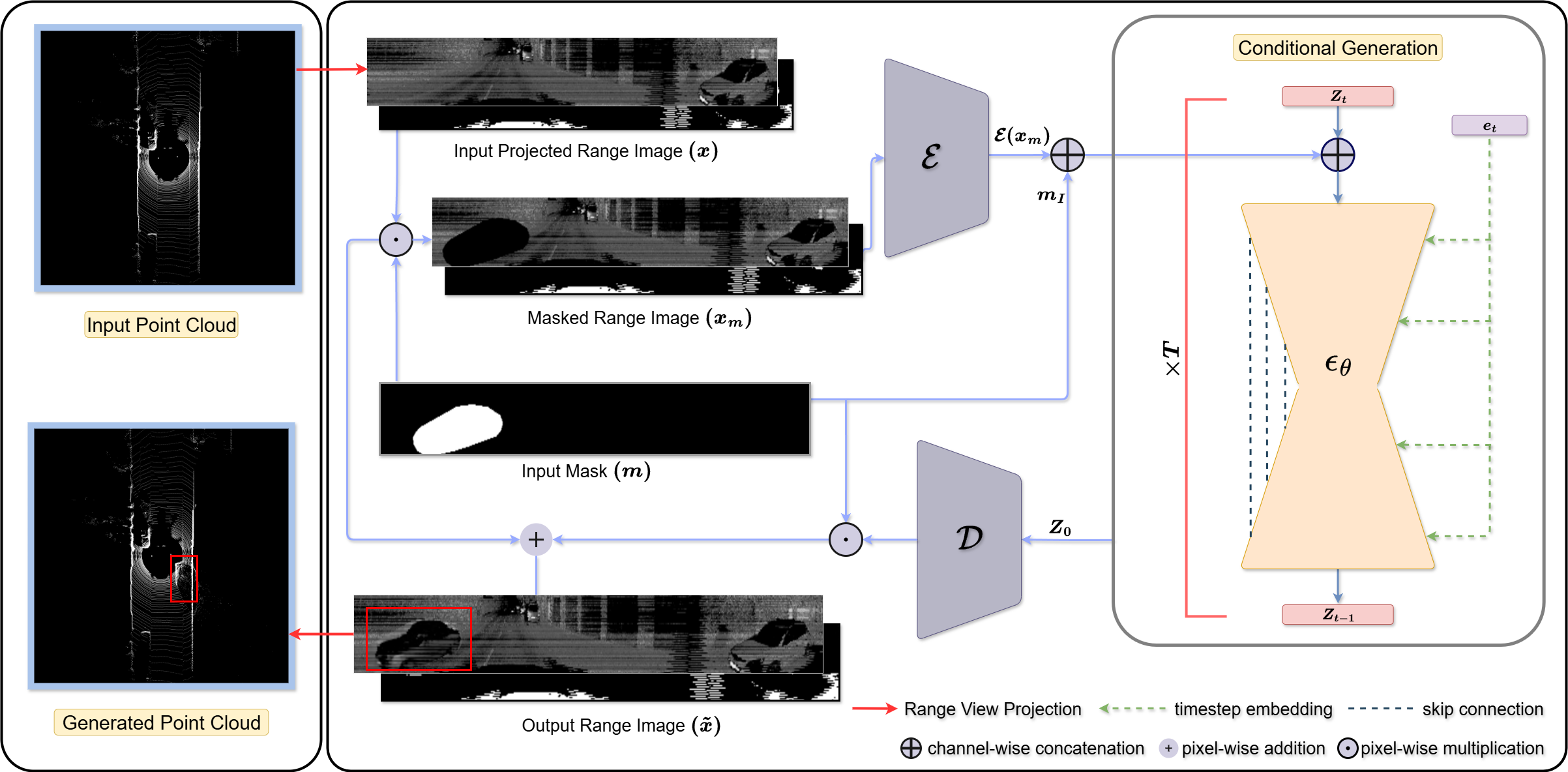}
    \caption{An Overview of the proposed method from the input point cloud to the generated output point cloud.}
    \label{fig:overview}
\end{figure*}

\subsection{Point Cloud Generation}
\textbf{Generative models for point clouds}.
In the past few years, a number of techniques have been explored and developed for point cloud generation, often inspired by existing methods in 2D image generation. 3D Point cloud generation methods have shown significant potential in the generation of outdoor and indoor environments. GANs have been one of the first models used to generate point clouds, with generator \textit{G} and discriminator \textit{D} jointly optimized with a mini-max objective \cite{achlioptas2018learningrepresentationsgenerativemodels,valsesia2018learning,shu20193dpointcloudgenerative}. VAEs have also found success in point cloud generation, they are generative models consisting of encoder and decoder networks that encode into and decode from a latent space capable of producing diverse samples \cite{kim2021setvaelearninghierarchicalcomposition,meng20253dpointcloudgeneration}. Normalizing flows are a class of probabilistic generative models that use a series of invertible and differentiable functions to transform between base and target distributions. They have also found relevance in point cloud generation \cite{yang2019pointflow3dpointcloud,klokov2020discretepointflownetworks,kim2020softflowprobabilisticframeworknormalizing}. Autoregressive models have also been employed by a few works \cite{sun2019pointgrowautoregressivelylearnedpoint,meng20253dpointcloudgeneration,li2023generalpointmodelautoencoding}, which work by sequentially generating points of a point cloud, usually using attention mechanisms. Over the past few years, diffusion models have found great success in image generation and, as a consequence, have inspired works to translate their success to 3-D generation too. \cite{zhou20213dshapegenerationcompletion,luo2021diffusionprobabilisticmodels3d,nichol2022pointegenerating3dpoint} are some of the works that have achieved diffusion-based point cloud generation.

\textbf{LiDAR scene generation}. The goal of LiDAR scene generation is to generate point clouds similar to those captured by a LiDAR scanner in real-world scenes. Unlike the generation of individual and isolated 3D models, scene generation poses additional challenges due to the scale and complexity of the scene and the realism expected from a generated point cloud. Earlier works \cite{caccia2019deepgenerativemodelinglidar,sallab2019lidarsensormodelingdata} used GANs as their generative model of choice, whereas \cite{caccia2019deepgenerativemodelinglidar,xiong2023ultralidarlearningcompactrepresentations} explored the use of variants of VAEs to achieve LiDAR generation. More recent works have shifted their focus to diffusion-based generative models, owing to their improved generation quality and greater flexibility in conditioning the generation process \cite{zyrianov2022learninggeneraterealisticlidar,hu2024rangeldmfastrealisticlidar,ran2024realisticscenegenerationlidar}.

\subsection{Data Representation}
Point clouds can be represented in different ways, such as range images, voxel grids, or BEV (birds-eye-view) projections, each offering trade-offs between spatial resolution, computational efficiency, and suitability for specific downstream tasks. Each representation has its own set of advantages and disadvantages \cite{zhou2018voxelnet,lang2019pointpillars,fan2021rangedet}. \textbf{Point-based methods} directly operate on raw point clouds, which make them computationally expensive but preserves the accuracy and inherent irregularity of point clouds and avoids the information loss that is associated with other representations \cite{shi2019pointrcnn, shi2020pointgnngraphneuralnetwork}. \textbf{Voxel-based methods} discretize 3D space into cubes and map points to these cubes. Although this discretization introduces quantization losses, voxel-based methods still found use in many convolution-based models \cite{deng2021voxelrcnnhighperformance,shi2022pillarnetrealtimehighperformancepillarbased,hu2024rangeldmfastrealisticlidar}. \textbf{Multi-perspective fusion methods} attempt to aggregate multiple representations, aiming to leverage the strengths and mitigate the weaknesses of each, though often at the expense of computational efficiency \cite{hu2022pointdensityawarevoxelslidar,xu2021rpvnetdeepefficientrangepointvoxel}. \textbf{Range-view image methods} exploit the LiDAR sensor sampling process to provide a representation unique to LiDAR data. This representation enables the conversion between 3D LiDAR data and 2D range-view images, facilitating the use of well-established 2D image-based techniques \cite{8967762,hu2024rangeldmfastrealisticlidar,ran2024realisticscenegenerationlidar}.

\subsection{Semantic Image Editing}
The field of semantic image editing covers a wide range of tasks, such as style transfer \cite{8953766}, inpainting \cite{9010928,li2022mat}, colorization \cite{2025PatRe.15911109C,9093389}, pose manipulation \cite{8237529}, and object manipulation (insertion, deletion, or translation) \cite{9577414,DBLP:conf/iclr/CouaironVSC23}. Semantic image editing excels at changing the content and narrative of an image by modifying the scene’s story, context, or elements. Tasks within this category include object addition, removal, replacement, and background modification \cite{10884879}. GANs \cite{ling2021editgan,10.5555/3495724.3496292,10.1145/3478513.3480538} and VAEs \cite{9010928,10.1007/978-3-030-58571-6_10,10204933} are widely used in earlier works that pioneered the task of semantic image editing. Scene graph manipulation-based methods \cite{9157808,10350451,10.1145/3687957} have found success in editing complex scenes where precise semantic relationships between objects are necessary. Diffusion models \cite{DBLP:conf/iclr/CouaironVSC23,10203581} also established themselves in works that traded higher computational requirements for high-fidelity image generation. Their ability to integrate multiple conditioning signals (text, class labels, images, semantic maps, image embeddings, etc.) made them the dominant approach in recent research. We attempt to translate this success to range-view images.

%% file: sec/method.tex
\section{Method}
\label{sec:method}
Figure~\ref{fig:overview} provides an overview of the proposed methodology. Initially, the point cloud is transformed into a 2D range view representation using the spherical projection technique, as described in the following subsection. The range view representation was chosen for this work because it allows the use of established 2D image editing techniques. Previous state-of-the-art models in LiDAR point cloud generation also demonstrated the advantages of using this representation \cite{hu2024rangeldmfastrealisticlidar,ran2024realisticscenegenerationlidar}. After obtaining the range-view representation, a diffusion-based image generation method was applied to perform the desired edits on the point cloud. Semantic masks served as a conditioning input during the generation process, the benefits of which are explained in Section 3.2. Finally, the generated range image is transformed back into a 3D LiDAR point cloud by performing the inverse of the original range projection.

\begin{table*}[h]
    \centering
    \begin{tabular}{|c|c|c|c|}
    \hline
       Model   & JSD ($\times10^{-2}$) $\downarrow$ & MMD ($\times10^{-3}$) $\downarrow$ & CD $\downarrow$\\
    \hline
       w/o convex hull & 1.880 & 24.685 &20.120 \\
       w/o region focused loss    & 1.766 & 12.156 &11.546 \\
        Ours   & 1.089 & 9.000 &8.351 \\
    \hline
    \end{tabular}
    \caption{Generated point cloud statistical metrics show that the model performance is poor when the convex hull is not computed during semantic masking and when the region focus loss is not implemented in training.}
    \label{tab: loss ablation}
\end{table*}

\begin{figure}[h]
    \centering
    \includegraphics[width=0.95\linewidth]{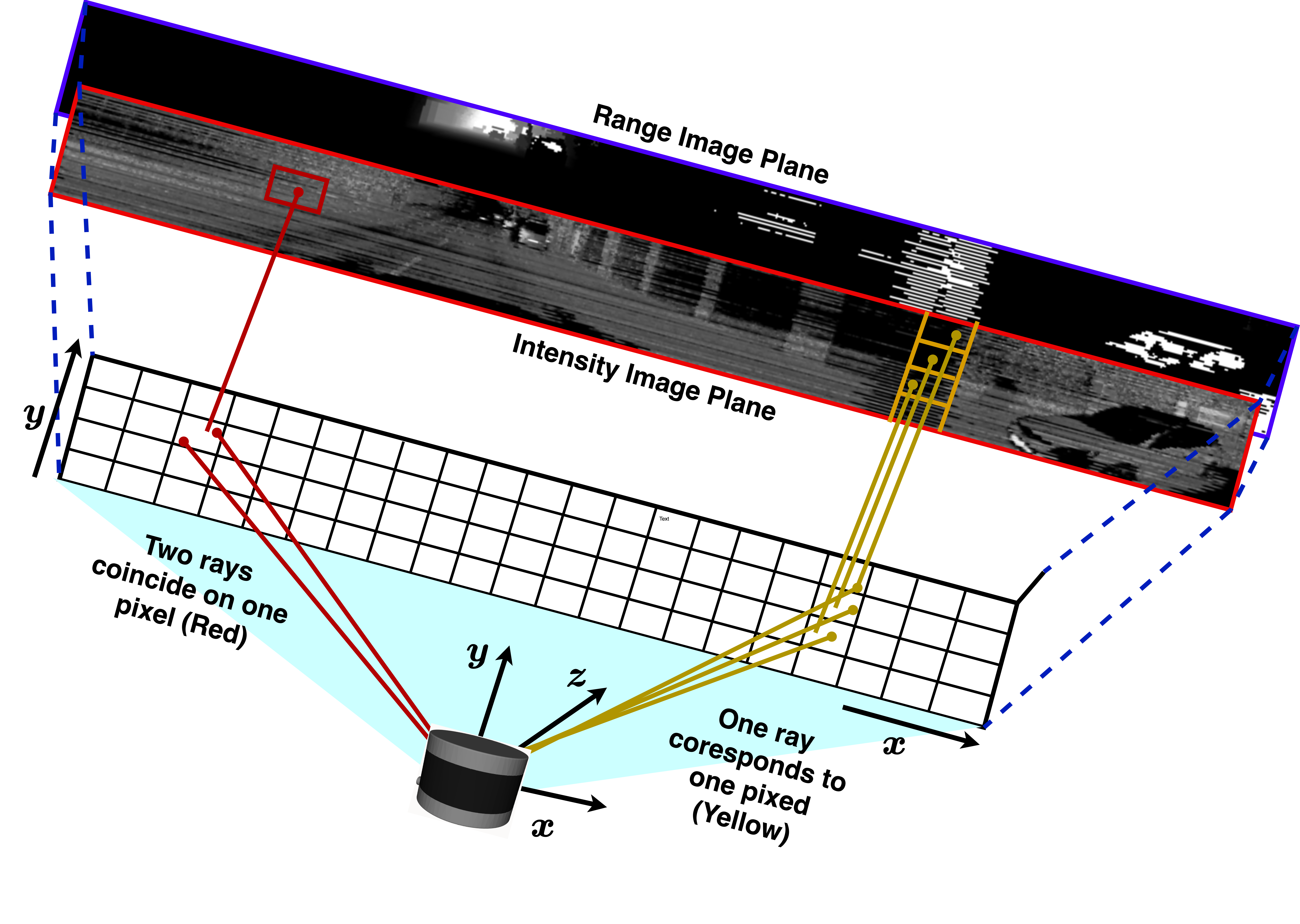}
    \caption{Range-View Projection}
    \label{fig:projection}
\end{figure}

\subsection{Range View Projection}
The range view projection is a 2D image representation of a LiDAR point cloud, denoted as \textit{$I_{r}$} $\in \mathbb{R}^{H \times W \times C}$, where \textit{H}, \textit{W}, and \textit{C} represented the height, width, and number of channels of the image, respectively. We follow the implementation of this projection as described in RangeLDM \cite{hu2024rangeldmfastrealisticlidar}. Each LiDAR point is represented as a 4-tuple $(x, y, z, i)$, where $x$, $y$, and $z$ denote the coordinates in a Cartesian coordinate system, and $i$ indicates the corresponding intensity value measured by the LiDAR scanner. The conversion from Cartesian coordinates $(x, y, z)$ to spherical coordinates $(r, \theta, \phi)$ is performed as follows, while considering the LiDAR scanner as the ego-center (origin) of the coordinate systems:
\begin{align}
r &= \sqrt{x^{2} + y^{2} + z^{2}} \\
\theta &= \operatorname{atan}(y, x) \\
\phi &= \operatorname{atan}\left(z, \sqrt{x^{2} + y^{2}}\right)
\end{align}
The points $(r,\theta,\phi)$ are rasterized onto an open cylinder with $W \times H$ grids to obtain the range-view image \textit{$I_{r}$} $\in \mathbb{R}^{H \times W \times 2}$, where $H$ and $W$ represent the height and width, respectively, and the two channels corresponded to the range and intensity values of the LiDAR points. Figure~\ref{fig:projection} shows an example of a range-view projection. The projection is invertible, enabling an easy transformation between the range-view image and the LiDAR point cloud. Once the range-view image is obtained, it is subjected to semantic masking.

\begin{figure}[h]
    \centering
    \includegraphics[width=0.95\linewidth]{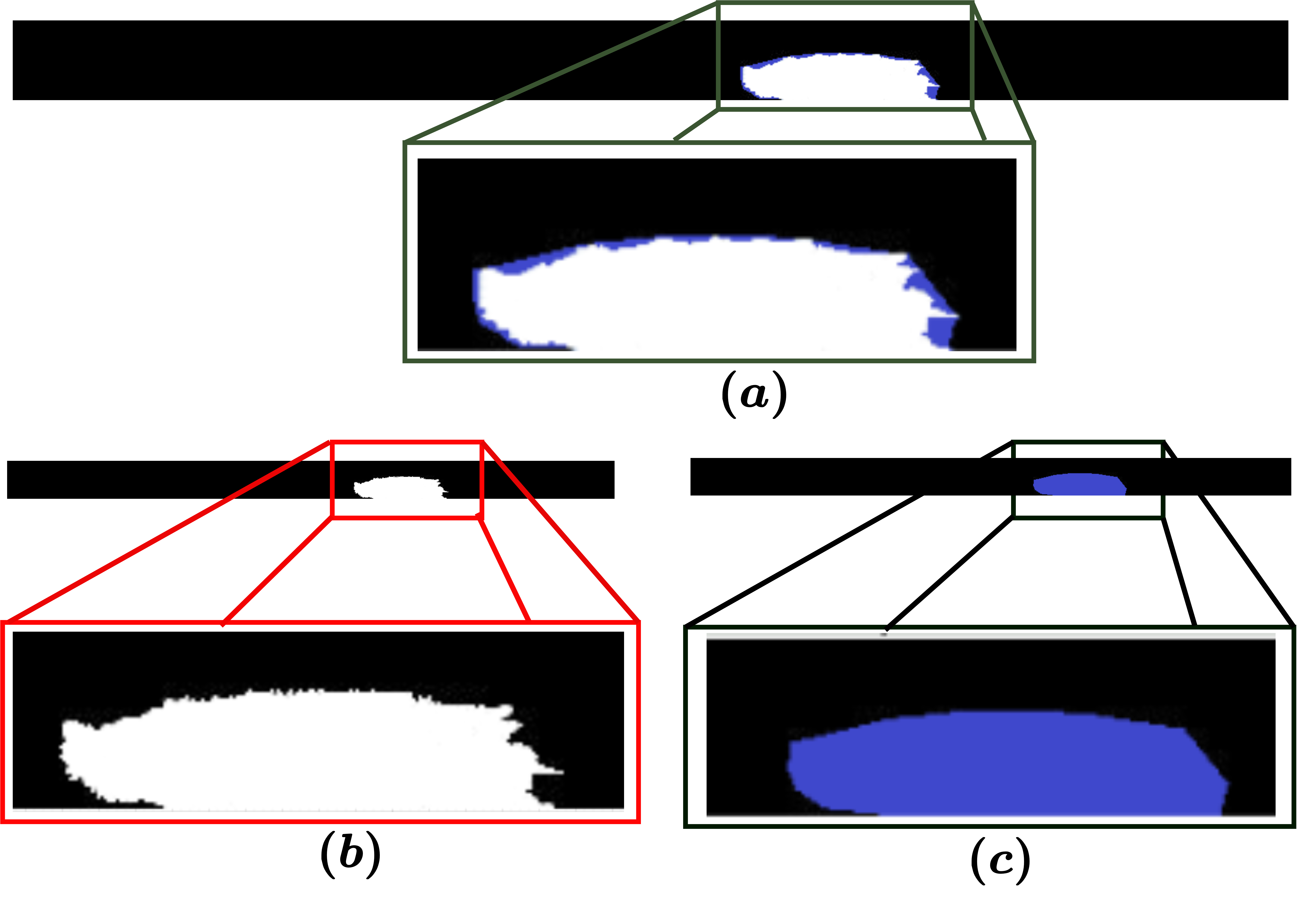}
    \caption{(a). Comparison between the mask made by the convex hull of the projected points and the mask of the projected points. (b). Mask formed from projected points. (c). Mask formed using a convex hull on the projected points.}
    \label{fig:sem_mask}
\end{figure}

\subsection{Convex Hull based Semantic Masking}
Point cloud editing poses major challenges: localizing the region of edit, asserting control over object dimensions and orientation within the scene, and ensuring that the generated content remains geometrically consistent with the surrounding region while simultaneously exhibiting a LiDAR-like distribution. We address some of these challenges using semantic masking. Within the range view image, we construct a semantic mask, as illustrated in Figure~\ref{fig:sem_mask}. This enabled us to define the desired object dimensions, orientation, and location information. We additionally compute the convex hull over the semantic mask to enhance geometric consistency during generation. The convex hull is computed over the set of pixels defining the mask, resulting in a convex polygon, as shown in Figure ~\ref{fig:sem_mask}. This convex polygon helps to regularize the shape of the object represented by the mask. Furthermore, it includes points surrounding the object within the region of interest, thereby forcing the generation process to incorporate these points as well. This approach improves both the geometric consistency and realism of the background of the generated output. The semantic mask then serves as a condition to the diffusion-based range view image generation process.

\begin{figure*}[h]
    \centering
    \includegraphics[width=0.95\linewidth]{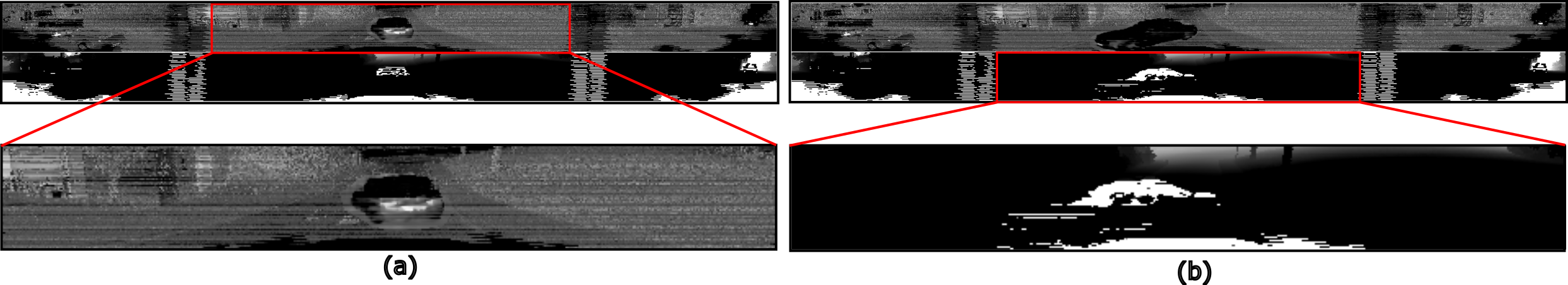}
    \caption{Qualitative results showing two examples of generated range-view images. (a). A close-up view of the intensity channel of the range-view image, where fine details like rear-light and license plate intensities are generated. (b). A close-up view of the range channel of the range-view image, where the ray-drop effect due to the windshield and windows is successfully generated.}
    \label{fig:range-view}
\end{figure*}

\subsection{Conditional Generation of Point Clouds}
The generation is carried out using an LDM (Latent Diffusion Model) \cite{rombach2022highresolutionimagesynthesislatent}, which consists of a VAE \cite{hu2024rangeldmfastrealisticlidar} and a Denoising Diffusion Probabilistic Model (DPDM) \cite{10.5555/3495724.3496298} operating within its constructed latent space of range view images. The pretrained VAE was kept frozen and is used to compress the range view images into a lower-dimensional latent space. This compression reduces the dimensionality of the data by encoding the most important features, thereby lowering the computational requirements while maintaining the quality of the range view images. DDMs are a class of generative models trained to invert the diffusion process. Over a number of timesteps T, the diffusion process gradually adds noise to the input data, modeled as a T-step Markov chain, until the resulting distribution becomes approximately Gaussian. The inverse of the diffusion process, referred to as the denoising or generative process, is achieved by training a neural network to iteratively predict the noise added across the T timesteps. The training of the neural network is performed by minimizing the following objective:
\begin{equation}
    L_{L D M}:={E}_{\mathcal{E}(x), \epsilon \sim \mathcal{N}(0,1), t}\left[\left\|\epsilon-\epsilon_{\theta}\left(z_{t}, e_{t}\right)\right\|_{2}^{2}\right]
\end{equation}
where $\mathcal{E}(x)$ is the encoded input range-view image $x$ using encoder $\mathcal{E}$ of the VAE, $\epsilon$ is the noise sampled from a Gaussian distribution at time step t. $\epsilon_{\theta}$ represents the noise estimator with parameters $\theta$, that is being trained to predict noise in a latent $z_{t}$ given time step embedding $e_{t}$ at time t.

Our adaptation of LDM modifies the standard objective described above to facilitate semantic mask guidance, which now reads:
\begin{align}
L_{LDM} :=\; &\mathbb{E}_{\mathcal{E}(x), m, \epsilon \sim \mathcal{N}(0, 1), t} \Big[ \nonumber \\
& \left\| \epsilon^{m_{I}} - \epsilon^{m_{I}}_{\theta}\left(z_{t}, e_{t}, \mathcal{E}(x_{m}), m_{I}\right) \right\|_{2}^{2} \Big]
\label{eq:ldm_loss}
\end{align}

The additional inputs to the neural network (UNet) include $\mathcal{E}(x_m)$, which represents the encoded masked input range view image, and $m_I$, which is the semantic mask interpolated to match the latent dimensions of the encoded image. The generation process is conditioned on the encoded masked input image ($\mathcal{E}(x_m)$) and the interpolated semantic mask ($m_I$), both of which are concatenated with $z_{t}$ at each timestep. The masked image serves to convey information around the masked region of the LiDAR point cloud. The interpolated mask accurately defines the masked region within the latents of the encoded range-view images; it is required despite having a mask region defined in the masked range-view image, as the pixels in the masked image became distorted by the encoder since the encoder has originally been trained to encode complete range-view images. These two conditioning factors guided the generation process by providing the necessary information for the edit.
Additionally, the $m_{I}$ in $\epsilon^{m_{I}}$ and $\epsilon^{m_{I}}{\theta}$ denote that the RMSE loss is calculated within the region specified by the interpolated mask $m_{I}$, as the generation is only focused on the masked region. This loss simplifies the objective and brings object-level focus during training.

%% file: sec/experiments.tex
\begin{figure*}[h]
    \centering
    \includegraphics[width=0.95\linewidth]{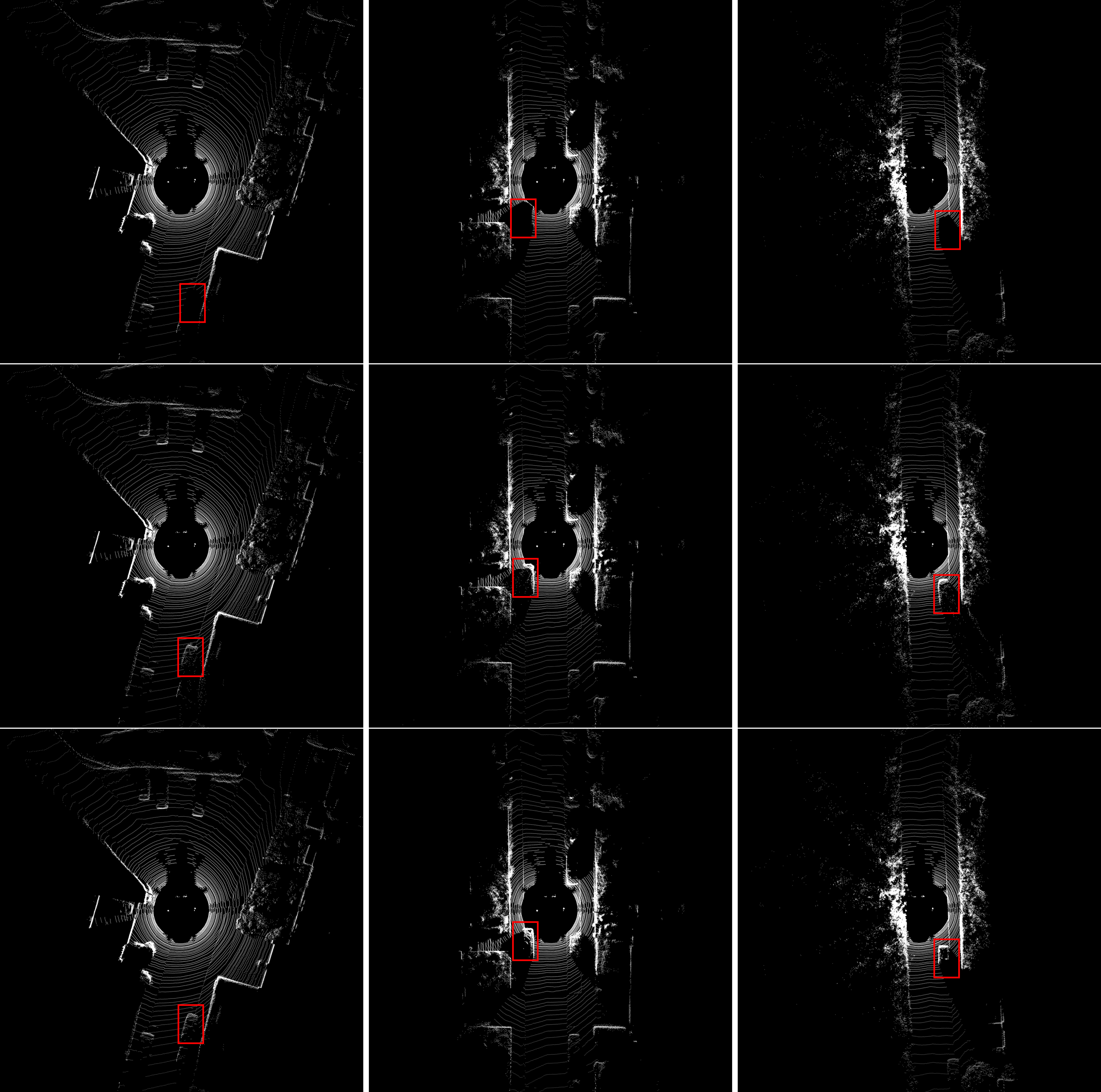}
    \caption{Qualitative results showing point cloud generation quality as BEV images on input masked point clouds. Masked point cloud (top row) vs generated point cloud (middle row) vs ground truth (bottom row).}
    \label{fig:gtcomparison}
\end{figure*}

\begin{figure*}
    \centering
    \includegraphics[width=0.95\linewidth]{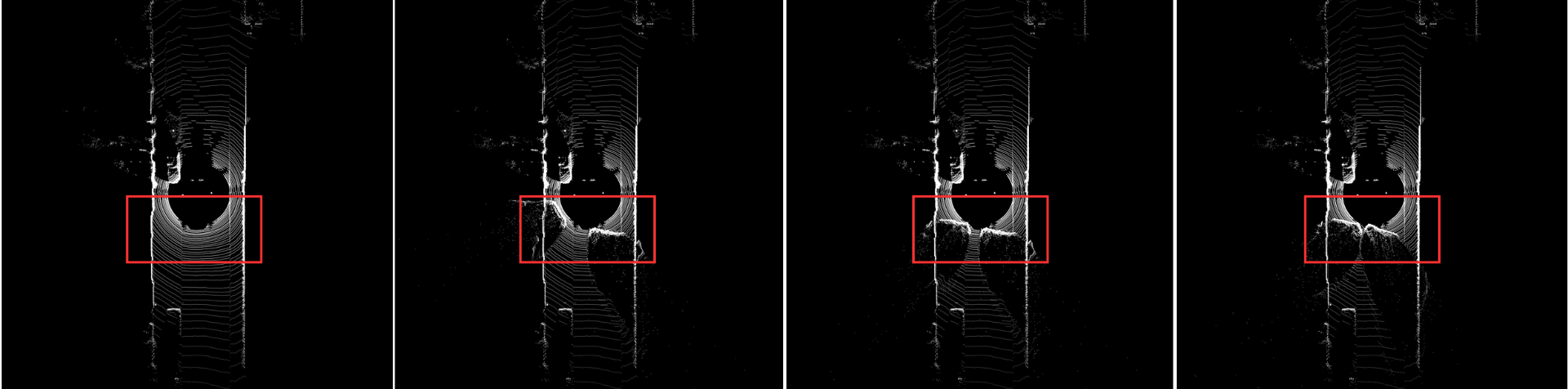}
    \caption{An input point cloud with a clear path in front of the ego vehicle (far left) transformed into a sequence of point clouds depicting a collision between two cars that have been generated in the path of the ego vehicle.}
    \label{fig:edge_case2}
\end{figure*}

\begin{figure*}[h]
    \centering
    \includegraphics[width=0.95\linewidth]{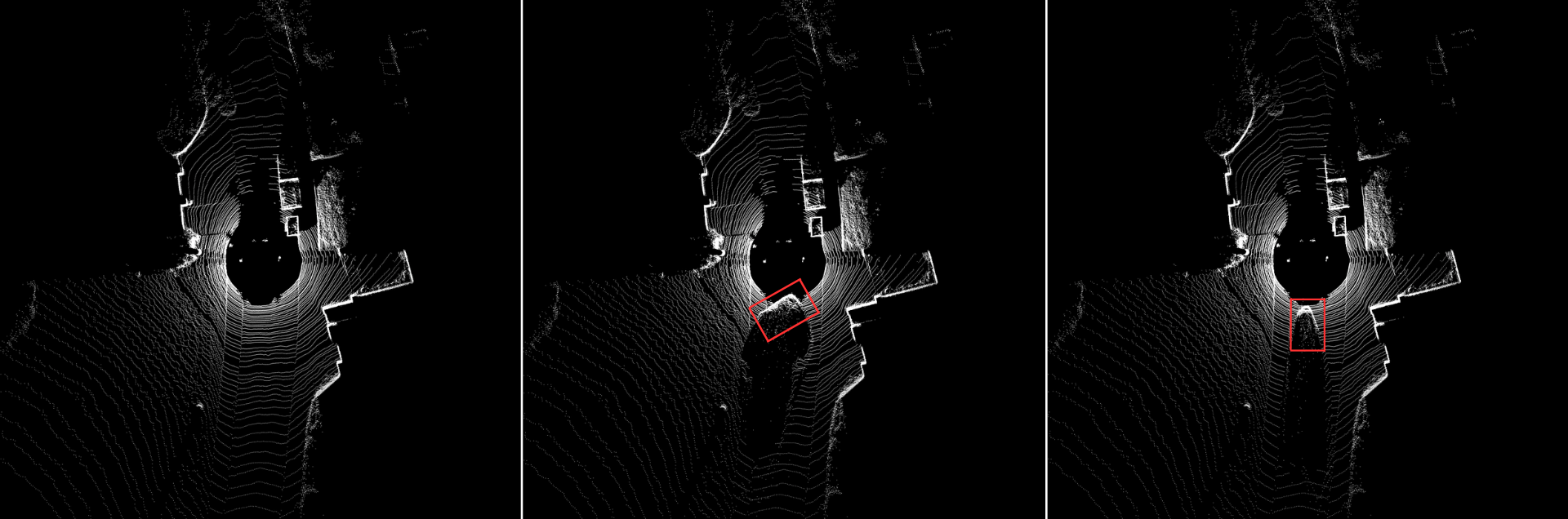}
    \caption{Original point cloud BEV image (left), generated point cloud BEV image (middle \& right) showing a potential edge case not present in the KITTI-360 dataset where the ego-vehicle path is obstructed by another car.}
    \label{fig:edge_case1}
\end{figure*}

\section{Experiments}
\label{sec:experiments}

\subsection{Experimental Settings}
All experiments are conducted on the KITTI-360 dataset \cite{geiger2012we}. We make use of the semantically-labeled point clouds and raw LiDAR scans to construct our training data; additional details of data creation, pre-processing, and training are provided in the supplementary. Our experiments are limited to a single object class, 'car', as this class is the most frequently occurring object class in the KITTI-360 dataset, which facilitates the diffusion models' need for a large number of training examples. Additionally, the object class 'car' is also of special interest for the point cloud editing task with respect to generating edge cases.

\subsection{Evaluation Metrics}
We evaluate the generation quality of the model at two levels: generated range-view images and generated point clouds. For judging the quality of generated range-view images, other works have used FID (also referred to as FRD or FRID) as a perceptual measure \cite{hu2024rangeldmfastrealisticlidar,ran2024realisticscenegenerationlidar}. This metric relies on comparing features obtained using the pretrained model RangeNet++ \cite{8967762}. Quality of generated point clouds has been evaluated by some of the previous works \cite{hu2024rangeldmfastrealisticlidar} using JSD (Jensen-Shannon Distance), MMD (Maximum Mean Discrepancy). We find LiDAR-EDIT's \cite{ho2025lidaredit} implementation relevant to our work as it allows for restriction of JSD and MMD calculation to specific object instances within the scene. We normalize the point clouds generated within the masked region to a unit sphere before the calculation of JSD and MMD metrics.

\begin{table}[h]
    \centering
    \begin{tabular}{|c|c|c|c|c|}
    \hline
       Model  & MAE $\downarrow$ & FRD $\downarrow$\\
    \hline
       w/o convex hull  & 0.313 & 969.159 \\
       w/o region focused loss & 0.285 & 965.889 \\
        Ours & 0.283 & 959.620 \\
    \hline
    \end{tabular}
    \caption{Generated range-view image metrics show that our addition of convex hull and region-focused loss has contributed to better FRD and MAE scores.}
    \label{tab: mask ablation}
\end{table}

\subsection{Effect of Convex Hull Masking}

The improvement provided by convex hull masking is captured quantitatively and shown in Tables~\ref{tab: mask ablation}, ~\ref{tab: loss ablation} in both the generated range-view images and generated point clouds, respectively. We attribute this improvement to the fact that the convex hull reduces the effects of noise in semantic point labels in the training data, helps overcome the irregularity in the shapes of real-world objects, and provides a general shape representation of objects in the range-view images.

\subsection{Effect of Region-Based Loss}
The use of region-based loss is done as our focus lies on object-level generation within the
real point clouds representing the complete scene, and thus, we only require the model training to focus on limited regions within the entire range-view images, as indicated by the mask. Region-based and other region-focused loss variants benefit the training of diffusion models in terms of convergence and improved quality in region-specific tasks like image editing. The advantage seen in our implementation is quantitatively represented in Tables~\ref{tab: loss ablation}, ~\ref{tab: mask ablation}.

\subsection{Generation Results}
The following sections will discuss the generation results obtained, showcasing the various capabilities and ease of use that our method brings in achieving LiDAR point cloud edits.

Our work simultaneously generates high-quality range and intensity information as part of the range and intensity channels of the range-view images, as shown in Figure~\ref{fig:range-view}. The model successfully generates fine-grained details as seen in the intensity image and also shows the ray-drop effect as expected of a real-LiDAR scan, without requiring any additional post-processing. 

\begin{figure}
    \centering
    \includegraphics[width=0.95\linewidth]{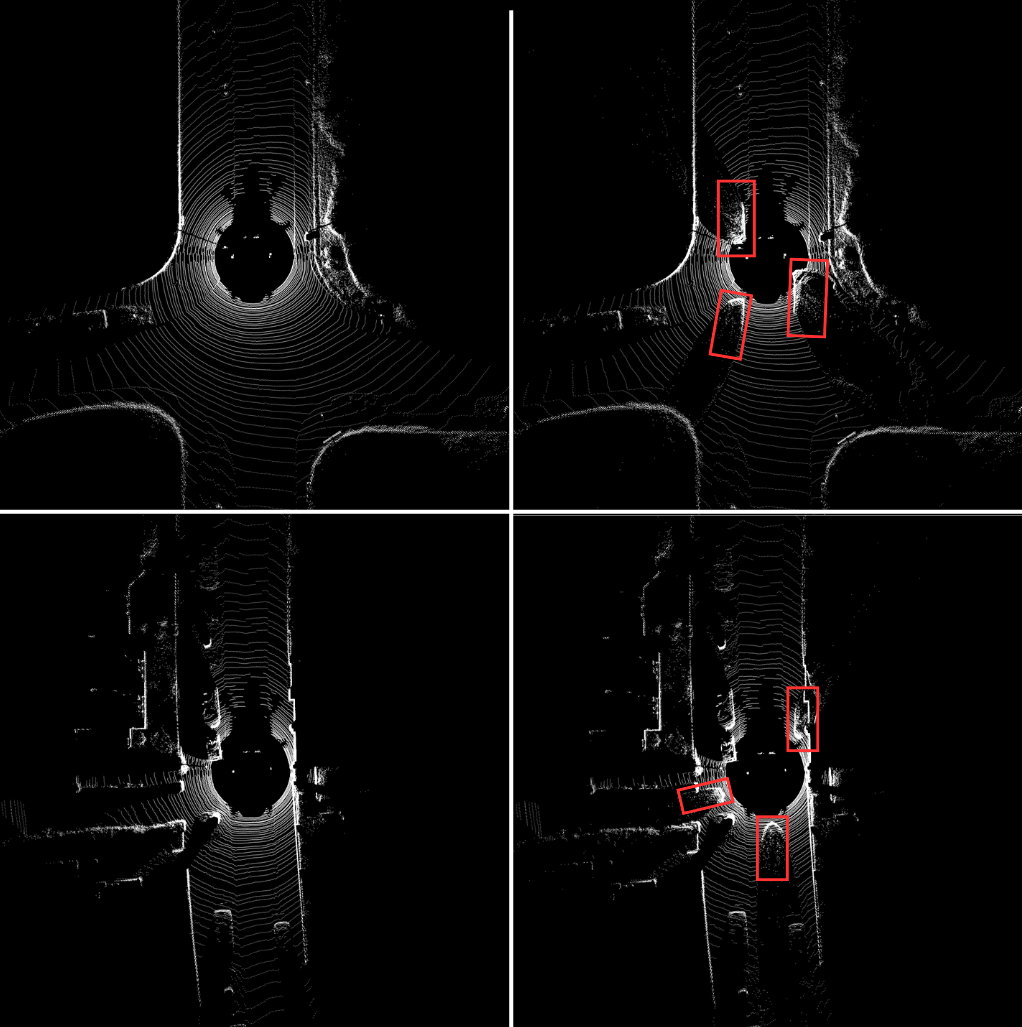}
    \caption{Original BEV image (left) and generated BEV image (generated cars in red box), where we have increased the number of cars present in the scene, simulating different traffic conditions.}
    \label{fig:multiobjects}
\end{figure}

 The generated LiDAR point clouds are close to real LiDAR point clouds in terms of quality and realism, as shown here in Figure~\ref {fig:gtcomparison}, a comparison is made between the ground truth and generated point clouds in the form of BEV images. While the model has been trained to generate one object corresponding to a single semantic mask at a time, it is easily extended to perform multiple object additions to the scene, as shown in Figure~\ref{fig:multiobjects} by repeating the generation process to sequentially add the objects as desired.

\begin{figure*}[h]
    \centering
    \includegraphics[width=0.95\linewidth]{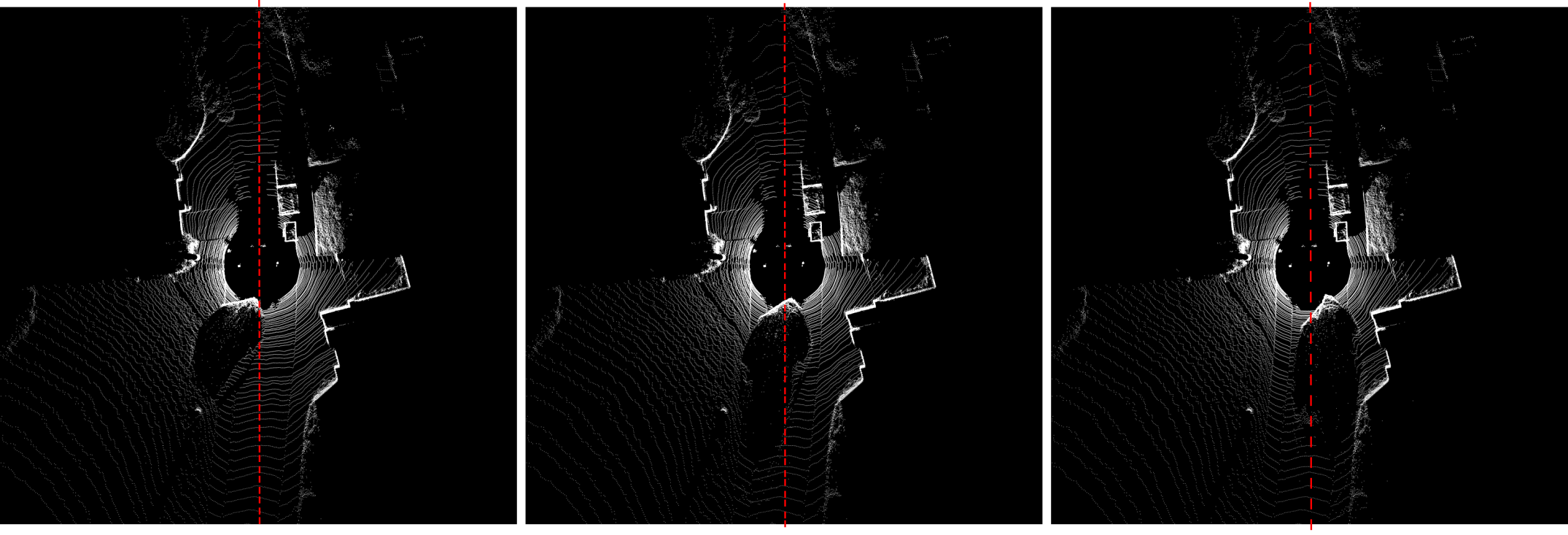}
    \caption{Example of dynamic object generation where the generated car appears to move across the reference \textcolor{red}{red} dotted line in front of the ego-vehicle.}
    \label{fig:edge_case3}
\end{figure*}

\subsection{Edge-Case Generation}
Edge cases are of extreme importance in autonomous driving, and our work offers a novel way to tackle edge case generation, as explained in this section. Any scenario that is absent or underrepresented in the training of an autonomous system can be considered an edge case. The KITTI-360 dataset used in this work is constructed from real-world LiDAR scans, and thus it lacks adverse events as they would occur infrequently in the real world. These edge cases cannot be neglected, as they are safety-critical; we attempt to generate some of the edge cases that arise in real-world scenarios from object interactions within the scene. Figure~\ref{fig:edge_case1} displays two examples where a car is deliberately generated in a way that diagonally and frontally blocks the path of the ego vehicle; the orientation and positioning of these cars are precisely and easily declared within the semantic mask. Figure ~\ref{fig:edge_case2} shows two cars generated into the scene and made to collide in front of the ego-vehicle, and Figure ~\ref{fig:edge_case3} depicts a car driving across the front of the ego-vehicle. These scenarios do not exist in the KITTI-360 dataset but were successfully generated using our method trained on the same dataset.

\subsection{Dynamic Scene Generation}
Dynamic scenes involve complex interactions between any number of objects and the environment. It is a challenge to exert precise control in the task of generating dynamic scenes, but our use of semantic masks and range-view images allows us to maintain pixel-level control in the range-view images during the generation of objects in the scene, coupled with the ability to sequentially make edits, enables us to assert control over the generated objects and make the scene appear dynamic when the edited frames are put together as shown in figures ~\ref{fig:edge_case2} and ~\ref{fig:edge_case3}.

%% file: sec/conclusion.tex
\section{Conclusion}
\label{sec:conclusion}
We propose a novel framework that tackles the task of semantic editing of LiDAR point clouds with precise geometric and positional control while preserving realism. We formulate the problem as semantic image editing of range-view images with diffusion-based image generation. We generate both range and intensity values of the point cloud, which is a step towards more realistic LiDAR editing. Our experiments on the KITTI-360 dataset also show the utility of our method in generating edge cases that are of great importance for autonomous systems. 

\textbf{Limitations \& Future Work}: While the use of range-view image representation has allowed us to formulate the problem as a semantic image editing task, it has limitations. Objects farther from the ego vehicle appear smaller in the range image, and correspondingly require the semantic mask to be smaller, which makes it harder to convey precise geometric information. Additionally, while diffusion models offer better generation quality, the amount of data required is a major hurdle for training on smaller datasets. This method can be extended to facilitate multi-class object generation as future work. Point cloud editing of dynamic scenes, precise editing for small-scale objects (those represented by few points within a large scene), and realism in generated point clouds are also some of the challenges left to be tackled by future work.